\documentclass[10pt,twocolumn,letterpaper]{article}

\usepackage{wacv}
\makeatletter
\@namedef{ver@everyshi.sty}{}
\makeatother
\usepackage{times}
\usepackage{epsfig}
\usepackage{graphicx}
\usepackage{amsmath}
\usepackage{amssymb}
\usepackage{booktabs}
\usepackage{algorithm}
\usepackage{tabu}
\usepackage{algpseudocode}
\usepackage{amsmath}
\usepackage{amssymb}
\usepackage{booktabs}
\usepackage{graphicx}
\usepackage{multirow}
\usepackage{subcaption}
\usepackage{bbm}
\usepackage{multirow}
\usepackage{tikz}
\usepackage{comment}
\usepackage{amsmath,amssymb} 
\usepackage{color}
\usepackage{enumitem}
\usepackage{adjustbox}
\usepackage{lineno}

\def\wacvPaperID{224} 

\wacvalgorithmstrack   

\wacvfinalcopy 

\ifwacvfinal
\usepackage[breaklinks=true,bookmarks=false]{hyperref}
\else
\usepackage[pagebackref=true,breaklinks=true,colorlinks,bookmarks=false]{hyperref}
\fi
\begin{document}

\title{Hyperspherical Quantization: Toward Smaller and More Accurate Models}
\author{Dan Liu$^1$,  Xi Chen$^1$,  Chen Ma$^2$, Xue Liu$^1$ \\
$^{1}$McGill University, $^{2}$City University of Hong Kong \\
\tt\small{dan.liu4@mail.mcgill.ca}, \\
 \tt\small \{xi.chen11,xue.liu\}@mcgill.ca, \\
 \tt\small{chenma@cityu.edu.hk}
}


\maketitle
\thispagestyle{empty}

\begin{abstract}
Model quantization enables the deployment of deep neural networks under resource-constrained devices.
Vector quantization aims at reducing the model size by indexing model weights with full-precision embeddings, \textit{i.e.}, codewords, while the index needs to be restored to 32-bit during computation. Binary and other low-precision quantization methods can reduce the model size up to 32$\times$, however, at the cost of a considerable accuracy drop. In this paper, we propose an efficient framework for ternary quantization to produce smaller and more accurate compressed models. 
By integrating hyperspherical learning, pruning and reinitialization, our proposed Hyperspherical Quantization (HQ) method reduces the cosine distance between the full-precision and ternary weights, thus reducing the bias of the straight-through gradient estimator during ternary quantization. Compared with existing work at similar compression levels ($\sim$30$\times$, $\sim$40$\times$), our method significantly improves the test accuracy and reduces the model size. 

\end{abstract}

\section{Introduction}

Despite promising results in real-world applications, deep neural network (DNN) models usually contain a large number of parameters, making them impossible to deploy on edge devices. 
A significant amount of research has been made to reduce the size and computational overhead of DNN models through quantization and pruning. 

Pruning brings high sparsity, but cannot take advantage of compression and acceleration without customized hardware \cite{guo2017software_hardware}. 
Cluster-based quantization, such as vector quantization and product quantization, remarkably reduces the model disk footprint \cite{stock2019killthebits,martinez_2020_pqf,cho2021dkm}, but its memory footprint is larger than that of the low-precision quantization method \cite{dai2021vectorquant,rastegari2016xnor,courbariaux2015binaryconnect,zhu2016ttq,li2016twn}, as the actual weight values involved in computation remain full-precision \cite{dai2021vectorquant}. Ultra-low-precision quantization, \textit{e.g.}, binary \cite{hubara2016binarized,courbariaux2016binarized,courbariaux2015binaryconnect}, ternary \cite{li2016twn,zhu2016ttq}, and 2-bit quantization \cite{zhou2016dorefa,choi2018pact,esser2019lsq}, has fast inference and low memory footprint \cite{rastegari2016xnor}, but it usually leads to a significant accuracy drop, due to the inaccurate weights \cite{gholami2021survey} and gradients \cite{yin2019understanding}. 

Gradually discretizing the weights can overcome such non-differentiability \cite{louizos2018relaxed,jang2016categorical,chung2016hierarchical}, \textit{i.e.}, reducing the discrepancy between the quantized weights in the forward pass and the full-precision weights in the backward pass.  
However, it only performs well with 4-bit (or higher) precision as the ultra-low bit quantizer may seriously damage the weight magnitude leading to unstable weights \cite{gholami2021survey}. Intuitively, ternary quantization barely affects the sign of the weights, making the direction of weight vectors \cite{salimans2016weight} changes relatively more stable than their magnitude. Recently, many studies \cite{liu2016large,liu2017deephyperspherical,liu2017sphereface,deng2019arcface,TimRDavidson2018HypersphericalVA,SungWooPark2019SphereGA,BeidiChen2020AngularVH} show that the angular information \cite{liu2021learning} preserves the key semantics in feature maps.

We propose hyperspherical quantization (HQ), a method combining pruning and reinitialization \cite{frankle2018lottery,zhou2019deconstructing}
to produce accurate ternary DNN models with a smaller memory/disk footprint.
We first pre-train a DNN model with a hyperspherical learning method \cite{liu2017deephyperspherical} to preserve the direction information \cite{salimans2016weight} of the model weights, then apply our proposed approach to push the full-precision weights close to their ternary counterparts, and lastly, we combine the straight-through estimator (STE) \cite{bengio2013estimating} with a gradually increased threshold to fulfill the ternary quantization process. Our main contributions are summarized as follows:
\begin{itemize}[topsep=0pt]
\item We demonstrate that simply integrating pruning and reinitialization can significantly reduce the impact of weight discrepancy caused by the ternary quantizer. We unify the pruning and quantization thresholds to one to further optimize the quantization process.

\item Our method significantly outperforms existing works in terms of the size-accuracy trade-off of DNN models. For example, on ImageNet, our method can compress a ResNet-18 model from 45 MB to 939 KB (48$\times$ compressed) while the accuracy is only 4\% lower than the original accuracy. It is the best result among the existing results (43$\times$, 6.4\% accuracy drop). 
\end{itemize}

\section{Related Work}
\subsection{Hyperspherical Learning}
Hyperspherical learning aims to study the impact of the direction \cite{salimans2016weight} of weight vectors on DNN models. \cite{salimans2016weight} discovers that detaching the model weight direction information from its magnitude can accelerate training. \cite{liu2017deephyperspherical} shows that the direction information of weight vectors, in contrast to weight magnitude, preserves useful semantic meanings in feature maps. \cite{liu2017sphereface,deng2019arcface,wang2018cosface} propose to apply regularization to angular representations on a hypersphere to enhance the model generalization ability in face recognition tasks.  \cite{liu2018decoupled,liu2017sphereface,deng2019arcface,TimRDavidson2018HypersphericalVA,SungWooPark2019SphereGA,BeidiChen2020AngularVH,WeiyangLiu2021OrthogonalOT} further study the empirical generalization ability of hyperspherical learning.
\subsection{Quantization}
Low-bit quantization methods convert float values of weights and activations to lower bit values \cite{courbariaux2015binaryconnect,courbariaux2016binarized,rastegari2016xnor,mcdonnell2018training}. These methods make it possible to substantially reduce the computational cost during CPU inference \cite{vanhoucke2011improving}. For example, binary quantization \cite{courbariaux2015binaryconnect,courbariaux2016binarized} compresses full-precision weights into a 1-bit representation, thus significantly reducing the memory footprint by 32$\times$. 

Clustering-based weight quantization, such as product quantization \cite{ge2013optimized} and vector quantization \cite{gong2014compressing,carreira2017model,stock2019killthebits} focus on optimizing the size-accuracy trade-off and can significantly compress the disk footprint by grouping weight values to a codebook. Common approaches cluster weights through k-means \cite{stock2019killthebits,wu2016quantized,son2018clustering,gong2014compressing} and further finetune the clustering center by minimizing the reconstruction error in the network \cite{fan2020training,stock2019killthebits,wu2016quantized}. The compression ratio and accuracy trade-off can be adjusted by changing the number of groups. \cite{han2015deep} applies k-means-based vector quantization with pruning and Huffman coding to further reduce the model size. However, the codebook usually consists of 32-bit float numbers \cite{stock2019killthebits,martinez_2020_pqf}, the memory footprint during computation is uncompressed. 

Some mixed-precision quantization methods overcome the shortcomings of low-bit and clustering based methods by means of reinforcement learning \cite{wang2019haq}, integer programming \cite{cai2020zeroq}, and differentiable neural architecture search \cite{wu2018mixed}, so as to apply different bit-widths in model weights to optimize inference time and model size. However, mixed-precision still cannot effectively compress the model size due to the use of 8 to 32-bit weights. Other mixed-precision quantization methods assign different bit widths to layer weights according to various measures, including hardware \cite{wang2019haq,yao2021hawq} and second-order information \cite{dong2019hawq,shen2020q}. 

\subsection{Pruning}
Pruning consists of structured and unstructured methods. It can greatly compress redundancy and maintain high accuracy. Unstructured pruning brings high sparsity, but cannot take advantage of acceleration without customized hardware \cite{guo2017software_hardware}. Only structured pruning methods can reduce the inference latency and are easier to accelerate \cite{he2017channel,li2016pruning} because the original weight structures of the model are preserved. Unstructured pruning uses criteria, such as gradient \cite{mozer1989using,lecun1990optimal}, and magnitude \cite{han2015deep,molchanov2017variational} information, to remove individual weights; structured pruning \cite{li2016pruning,hu2016network,alvarez2017compression} aims to remove unimportant channels of the convolutional layer based on similar criteria. The lottery ticket hypothesis \cite{frankle2018lottery} shows that there exists sparse subnetworks that can be trained from scratch and achieve the same performance as the full network. \cite{zhou2019deconstructing} studies the lottery ticket hypothesis from the perspective of weight reinitialization and points out that the key premise is the sign of weight values.

Re-training after pruning \cite{frankle2018lottery,zhou2019deconstructing} reveals the link between the network structure and performance. Furthermore, our findings show that training after pruning and reinitialization can be used to produce more accurate and highly compressed ternary weights, which surpasses the current model compression methods and has a wide range of application scenarios.
\section{Preliminary and Notations}
\begin{figure*}
\centering
\includegraphics[width=0.8\textwidth]{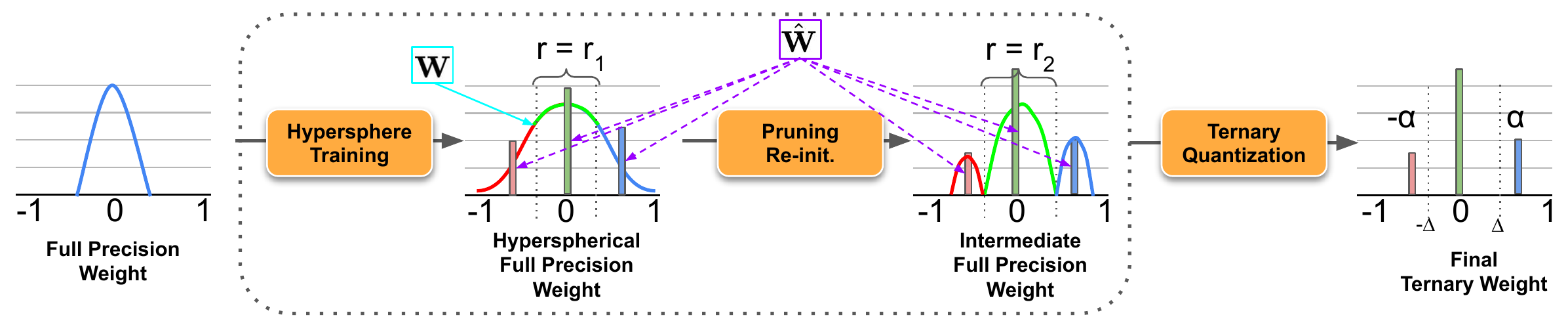} 
\caption{
The overall preprocessing and ternary quantization process of our proposed method (Section \ref{increase_s}). $\alpha={\frac{1}{\sqrt{|\mathbf{I}_\Delta|}}}$, $r_1=r_l$, and $r_2=r_h$. The dotted-line square denotes the preprocessing step. The ``\textbf{Pruning Re-init.}'' process can increase $\mathcal{S}(\mathbf{\hat{W}},\mathbf{{W}})$, \textit{i.e.}, reduce $\mathcal{D}$ (Section \ref{sec:Pstep}, \ref{Rstep}). We seek to minimize $\mathcal{D}$ before ``\textbf{Ternary Quantization}''. 
}
\label{fig:whole}
\end{figure*}
\subsection{\textcolor{black}{Hyperspherical Model}}
A general representation of a hyperspherical neural network layer \cite{liu2017deephyperspherical} is:
\begin{equation}
\label{eq:spop}
    \mathbf{y}=\phi(\mathbf{W}^T\mathbf{x}),
\end{equation}
where $\mathbf{W} \in \mathbb{R}^{n\times{m}}$ is the weight matrix, $\mathbf{x}\in \mathbb{R}^{n}$ is the input vector to the layer, $\phi$ represents a nonlinear
activation function, and $\mathbf{y}\in \mathbb{R}^{m}$ is the output feature vector. The input vector $\mathbf{x}$ and each column vector $\mathbf{w}_j\in\mathbb{R}^{n}$ of $\mathbf{W}$
satisfy $\|\mathbf{w}_j\|_2=1$,$\|\mathbf{x}\|_2=1$ for all $j = 1, . . . , m$.
\subsection{\textcolor{black}{Ternary Quantizer}}
In this work, the ternary quantizer is:
\begin{equation}
\label{eq_sign}
\hat{\mathbf{W}}=\texttt{Ternary}(\mathbf {W},\Delta)=\left\{\begin{aligned} {\frac{1}{\sqrt{|\mathbf{I}_\Delta|}}} &: {w_{ij}}>\Delta, \\ 0 &:\left|{w_{ij}}\right| \leq \Delta, \\-{\frac{1}{\sqrt{|\mathbf{I}_\Delta|}}} &: {w_{ij}}<-\Delta, \end{aligned}\right.
\end{equation}
where $\mathbf{W}$ is the full-precision weights, $\Delta$ is a threshold, $\mathbf{I}_\Delta=\{i||w_{ij}|>\Delta\}$  \cite{li2016twn}, and $|\mathbf{I}_\Delta|$ denotes total non-zero values in the $j$-th column vector $\mathbf{w}_j$. With $\Delta=0$, $\texttt{Ternary}(\cdot)$ becomes a variant of $\texttt{Binary}(\cdot)$ operation. And
$   \label{hcode}
    \phi(\mathbf{\hat{w}}_j^T\mathbf{x})=\phi(\frac{1}{\sqrt{|\mathbf{I}_\Delta|}}\mathbf{\bar{w}}_j^T\mathbf{x})
$
s.t. $\mathbf{\bar{w}}\in \textcolor{black}{\{-1,0,1\}}$.
\subsection{\textcolor{black}{Pruning}}
The unstructured pruning \cite{han2015deep} is defined by:
\begin{equation}
\label{pruning}
\mathbf{W}'=\texttt{Prune}(\mathbf{W},r)=\mathbf{W}\odot\mathbf{M},
\end{equation}
where $\odot$ denotes the element-wise multiplication. Mask $\mathbf{M}$ selects the top $r$ percent of the smaller weights in $\mathbf{W}$:
\begin{equation}
     \mathbf{M}=\texttt{Mask}(\mathbf{W},r)=\texttt{Sign}(|\texttt{Ternary}(\mathbf {W},\Delta)|),
\end{equation}
where $\Delta=\texttt{threshold}(\mathbf{W}, r)$, and $0\le r < 1$. The $\texttt{threshold}(\mathbf{W}, r)$ returns the corresponding minimum value of matrix $\mathbf{W}$ based on the percentage $r$.
We use pruning to represent unstructured pruning for simplicity.

\subsection{\textcolor{black}{Cosine Similarity $\mathcal{S}$} on HyperSphere}
Based on Eq.~\eqref{eq_sign} and hyperspherical learning, we have $\|\mathbf{\hat{w}}_j\|_2=1$ and $\|\mathbf{{w}}_j\|_2=1$. The vector-wise cosine similarity between $\mathbf{w}_j$ and $\mathbf{\hat{w}}_j$ is:
\begin{equation}
    \label{sc_0}
    \mathcal{S}(\mathbf{\hat{w}}_j,\mathbf{{w}}_j)=\frac{\mathbf{\hat{w}}_j\cdot\mathbf{{w}}_j}{\|\mathbf{\hat{w}}_j\|_2\|\mathbf{{w}}_j\|_2} \,
    ={\sum_{i=1}^{n} \frac{1}{\sqrt{|\mathbf{I}_\Delta|}}|w_{ij}|}.
\end{equation}
If without pruning, $\mathbf{W}$ and $\mathbf{\hat{W}}$ will not contain $0$. With $\Delta=0$, we have $|\mathbf{I}_\Delta|=n$ and the cosine similarity between $\mathbf{\hat{W}}$ and $\mathbf{{W}}$ becomes:
\begin{equation}
    \label{sc_1}
    \mathcal{S}(\mathbf{\hat{W}},\mathbf{{W}})={\frac{1}{m}\sum_{j=1}^{m}\sum_{i=1}^{n} \frac{1}{\sqrt{n}}|w_{ij}|}.
\end{equation}
After applying pruning (Eq.~\eqref{pruning}) to $\mathbf{W}$:
\begin{equation}
    \label{sc}
    \mathcal{S}(\mathbf{\hat{W}}',\mathbf{{W}}')={\frac{1}{m}\sum_{j=1}^{m}\sum_{i=1}^{n} \frac{1}{\sqrt{|\mathbf{I}_\Delta|}}|w_{ij}'|}.
\end{equation}

The cosine distance between full-precision and ternary weights is:
\begin{equation}
\label{cosdist}
    \mathcal{D}=1-\frac{1}{l}\sum_{k=1}^{l}\mathcal{S}(\mathbf{\hat{W}}_k,\mathbf{{W}}_k),
\end{equation}
where $l$ denotes the number of quantized layers.

\section{\textcolor{black}{Hyperspherical Quantization}}\label{sec4}

In this section, we propose using pruning to increase the cosine similarity $\mathcal{S}$ between the full-precision weights $\mathbf{{W}}$ and the ternary weights $\mathbf{\hat{W}}$. We show how we can effectively quantize such discrepancy reduced model weights.

Our proposed method includes the preprocessing and quantization steps (Fig. \ref{fig:whole}). In the preprocessing step, we use iterative pruning with gradually increasing sparsity and reinitialization \cite{zhou2019deconstructing} to push $\mathbf{{W}}$ close to its ternary counterpart $\mathbf{\hat{W}}$. In the quantization step, as $\mathbf{W}$ is close to $\mathbf{\hat{W}}$ after the first step, it is easy to obtain a more accurate $\mathbf{\hat{W}}$ by using regular STE-based ternary quantization methods. We unify the thresholds of pruning and quantization as one single threshold during ternary quantization process. 


\subsection{\textcolor{black}{Increasing $\mathcal{S}$ by Preprocessing}}
\label{increase_s}
We show that pruning on hypersphere can increase the cosine similarity $\mathcal{S}(\mathbf{\hat{W}},\mathbf{{W}})$, thus pushing the full-precision weight close to its ternary version. But the decayed weights during training cause unstable $\mathcal{S}$, which makes ternary quantization infeasible. Then the reinitialization is proposed to stabilize $\mathcal{S}$.
\begin{figure*}[t]
     \centering
     \begin{subfigure}[b]{0.3\textwidth}
         \centering
         \includegraphics[width=\textwidth]{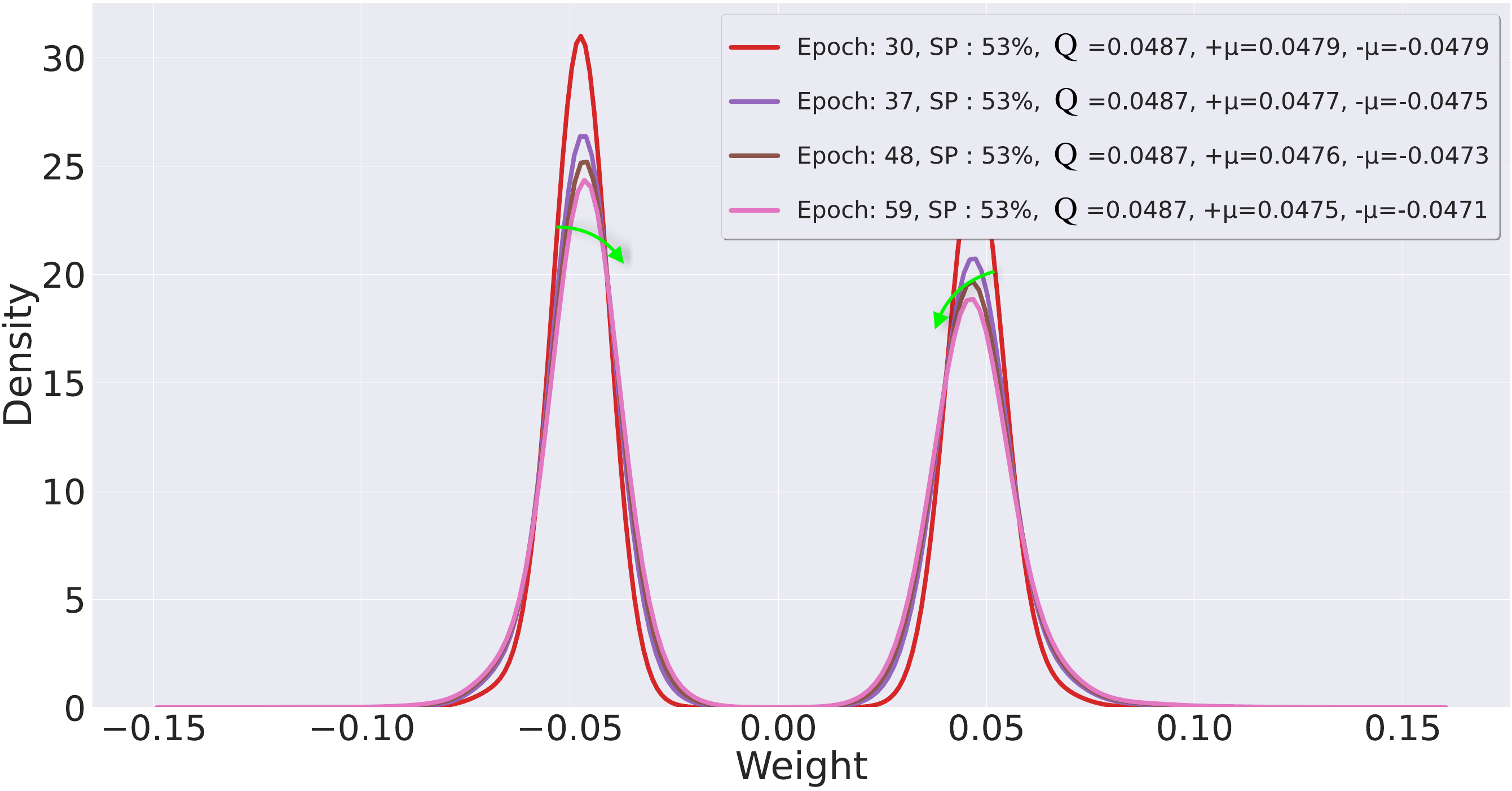}
         \caption{ResNet-50 with the fixed sparsity.}
         \label{fig:spfix}
     \end{subfigure}
     \begin{subfigure}[b]{0.3\textwidth}
         \centering
         \includegraphics[width=\textwidth]{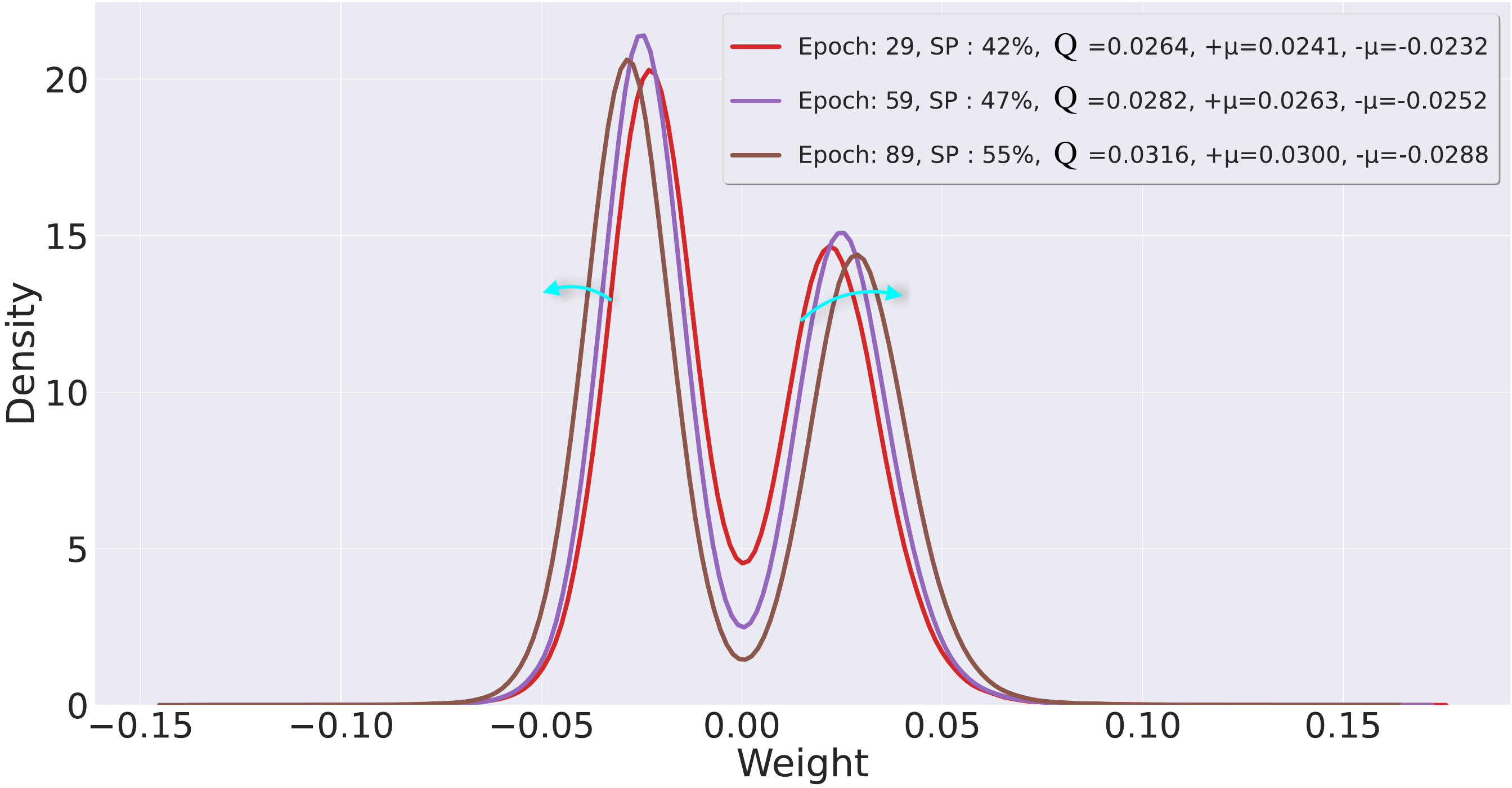}
         \caption{ResNet-50 with iterative processing.}
         \label{fig:sp_nfix}
     \end{subfigure}
    \caption{\textcolor{black}{ Weight distribution of a hidden convolutional layer. The zero-drifting is relieved by iterative pruning and reinitialization.  $Q=|\hat{w}|=|\frac{1}{\sqrt{\mathbf{I}_\Delta}}|$, $\pm\mu$ denotes the mean values of $w^+\in\mathbb{R}_{>0}$ and $w^-\in\mathbb{R}_{<0}$ during training.
    (a) When the sparsity is fixed, due to the decayed weights, $|\mu|$ is heading towards zero. (b) When we iteratively performe pruning and reinitialization,  $|\mu|$ grows with the sparsity and keeps moving closer to $|\hat{w}|$ as epoch increases.}}
    \label{img:prprpr}
\end{figure*}
\subsubsection{\textcolor{black}{ Cosine Similarity and Hyperspherical Pruning}}
\label{sec:Pstep}
Given a full-precision $\mathbf{W}$ and its ternary form $\mathbf{\hat{W}}$, we seek to optimize the following problem under hyperspherical learning settings:
\begin{equation}
    \max_{r}~~\mathcal{S}(\mathbf{\hat{W}}',\mathbf{{W}}')
\end{equation}
\begin{equation*}
    s.t.~~ 0<\mathcal{S}\le1,
\end{equation*}
where $\mathbf{W}'=\texttt{Prune}( \mathbf{W},r)$ and $\mathbf{\hat{W}}'=\texttt{Ternary}(\mathbf{W}',0)$. Obviously, if $\|\mathbf{W}'\|_0=1$ then $\mathcal{S}(\mathbf{\hat{W}}',\mathbf{{W}}')=1$, but it is meaningless.
Although there is no explicit solution for $r$, we can increase $\mathcal{S}$ by gradually increasing $r$.
Since the model is trained with hyperspherical learning, based on Eq.~\eqref{sc_1}-\eqref{sc}, with pruning ratio $r$, we have $\frac{1}{\sqrt{|\mathbf{I}_\Delta|}}\ge \frac{1}{\sqrt{n}}$ and:
\begin{equation}
    \label{w_pr}
    |w_{ij}'|=\frac{|w_{ij}|}{\|\mathbf{w'}\|_2} \ge \frac{|w_{ij}|}{\|\mathbf{w}\|_2}=|w_{ij}|,
\end{equation}
where ${\|\mathbf{w'}\|_2} \le {\|\mathbf{w}\|_2=1}$. Therefore:
\begin{equation}
    \label{w_pr_1}
    \mathcal{S}(\mathbf{\hat{W}}',\mathbf{{W}}')\ge\mathcal{S}(\mathbf{\hat{W}},\mathbf{{W}})
\end{equation}

Eq.~\eqref{w_pr}-\eqref{w_pr_1} indicate that pruning with ratio $r$ can increase the cosine similarity $\mathcal{S}(\mathbf{\hat{W}}',\mathbf{W}')$.
However, we still need a proper pruning ratio to maintain the model's performance,  as the over-pruned weight has a higher $\mathcal{S}$ but may not perform well. For example, $\mathbf{w}=[0.3, 0.2, 0.0001]$, $\mathbf{w}'=[0.3, 0.2, 0]$, $\hat{\mathbf{w}}=[1,1,1]$, and $\hat{\mathbf{w}}'=[1,1,0]$, then $\mathcal{S}(\hat{\mathbf{w}}',\mathbf{w}')=0.98$ is greater than $\mathcal{S}(\hat{\mathbf{w}},\mathbf{w})=0.80$. If we further apply pruning to $\mathbf{w}'$ then we have  $\mathbf{w}''=[0.3,0,0]$, $\mathbf{\hat{w}}''=[1,0,0]$, and  $\mathcal{S}(\hat{\mathbf{w}}'',\mathbf{w}'')=1.0$. It is obvious that $\mathbf{\hat{w}}''$ cannot perform as well as $\mathbf{\hat{w}}'$.

\subsubsection{\textcolor{black}{Zero-Drifting and Iterative Reinitialization}}
\label{Rstep}
In addition to the pruning ratio $r$, Eq.~\eqref{sc_0} indicates that $\mathcal{S}$ is also related to $|w_{ij}|$. In this section, we discuss how the decayed weights has negative impact on $\mathcal{S}$, and how to mitigate such impact through periodical reinitialization.

Weight decay and learning rate push the weight values toward and around zeros \cite{lr_lambda,nowlan1992simplifying,hinton1986learning}. Learning rate plays a similar role as weight decay (see Eq.~(8) in the work of \cite{hanson1988comparing}). As $w_{ij}$ drifting toward zero during training (Fig.~\ref{fig:spfix}), $\mathcal{S}(\mathbf{\hat{W}},\mathbf{{W}})$ is also reduced (Eq.~\eqref{sc_0}). Intuitively, this zero-drifting phenomenon hinders accurate ternary quantization as the reduced $\mathcal{S}(\mathbf{\hat{W}},\mathbf{{W}})$ enlarges the quantization searching space.
We reinitialize the weights by $\mathbf{W}=\texttt{Ternary}(\mathbf{W},0)$ to neutralize the negative effect of the decayed weights.

Fig.~\ref{fig:spfix} shows the zero-drifting phenomenon with fixed pruning sparsity. Our experimental results show that iterative pruning and reinitializing can increase $\mathcal{S}(\mathbf{\hat{W}},\mathbf{{W}})$ and prevent the negative impact of decayed weights (Table \ref{tab_res_quant_distance},\ref{tab_coco_quant_distance}).  Once we periodically prune, reinitialize and re-train the model, we always have a larger $\mathcal{S}$ (Table \ref{tab_res_quant_distance},\ref{tab_coco_quant_distance}) and 
the zero-drifting is relieved (Fig.~\ref{fig:sp_nfix}, Table \ref{tab_res_quant_distance}).

\subsection{\textcolor{black}{Quantization with a Unified Threshold}}
\label{subsec:WQ}
With $\mathbf{W}$ close to $\hat{\mathbf{W}}$ in the preprocessing step, we still need to perform weight quantization and pruning to further increase $\mathcal{S}(\hat{\mathbf{W}}, \mathbf{W})$. We introduce a gradually increasing quantization threshold $\Delta$ (Eq. \ref{eq_sign}) to unify pruning and quantization, as $\Delta$ can be seen as a pruning threshold. The non-differentiability of $\texttt{Ternary}(\cdot)$ is bypassed with STE \cite{bengio2013estimating}:

\textbf{Forward:}
\begin{align}
\begin{split}
\hat{\mathbf{W}}={\texttt{Ternary}(\mathbf{W},\Delta)}
\end{split}
\end{align}

\textbf{Backward:}
\begin{equation}
    \label{eq9}
    \frac{\partial E}{\partial {\mathbf{W}}}=\frac{\partial E }{\partial {\hat{\mathbf{W}}}} \frac{\partial \hat{\mathbf{W}} }{\partial {\mathbf{W}}}\underset{S \bar{T} E}{\approx} \frac{\partial E }{\partial {\hat{\mathbf{W}}}}.
\end{equation}
STE essentially ignores the quantization operation and approximates it with an identity function. 
\begin{figure*}
     \centering
     \begin{subfigure}[b]{0.3\textwidth}
         \centering
         \includegraphics[width=\textwidth]{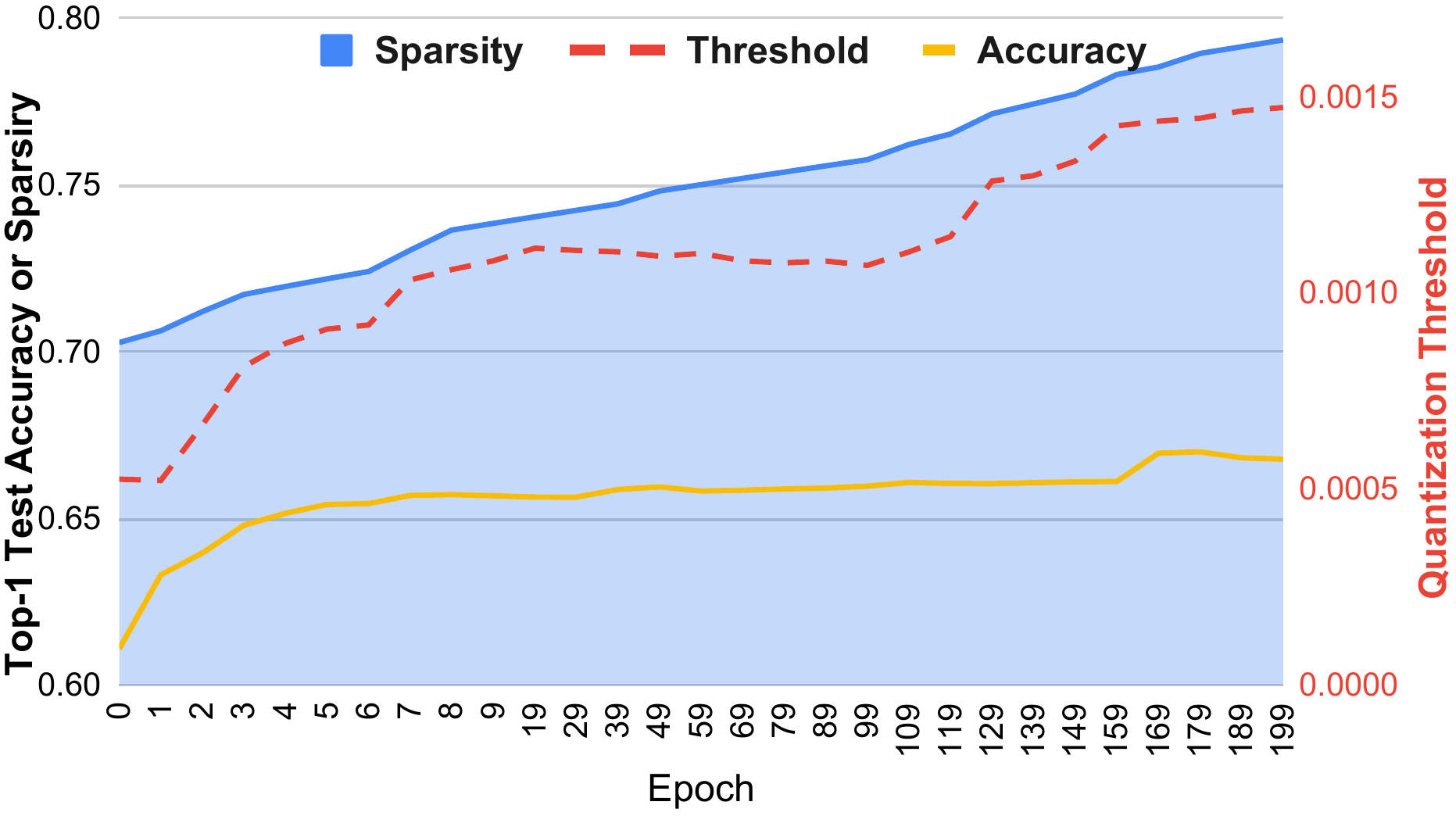}
         \caption {ResNet-18 on ImageNet.}
         \label{fig:y equals x}
     \end{subfigure}
     \begin{subfigure}[b]{0.3\textwidth}
         \centering
         \includegraphics[width=\textwidth]{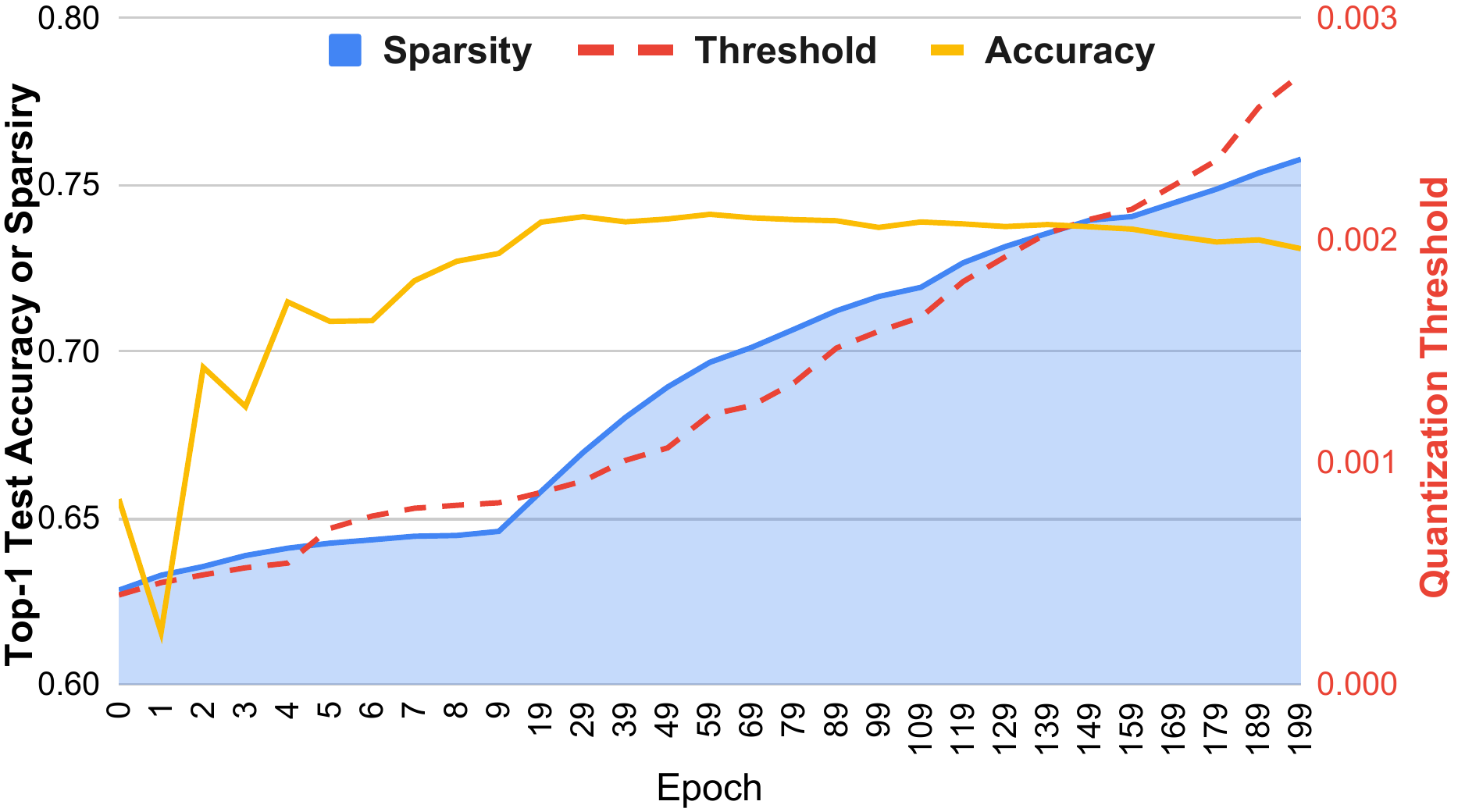}
         \caption {ResNet-50 on ImageNet.}
         \label{fig:three sin x}
     \end{subfigure}
    \caption{The trend of sparsity, threshold $\Delta$ and accuracy during ternary quantization. The blue area denotes model sparsity.}
    \label{img:threshodline}
\end{figure*}

The threshold $\Delta$ should gradually increase along with the training error, and such increase should slow down after the model converges (Fig.~\ref{img:threshodline}). Therefore, we directly use the averaged gradients to update the pruning threshold:
\begin{equation}
    \mathbf{\Delta} =\sum_{i=1}^{n}{\frac{\partial L}{\partial w_{i}}}, w_{i}\in \mathbf{w}  \text{ and } w_{i} \ne 0 .\label{eq19}
\end{equation}
The gradient of $\frac{\partial L}{\partial w_{i}}$ where $w_{i}=0$ is ignored. The training error is very large at the beginning, leading to a rapid increase of the threshold. As the model gradually converges, the training error will decrease and the threshold growth slows down (Fig.~\ref{img:threshodline}). The model sparsity is gradually increasing along with the learned threshold. The accuracy starts to decrease if the quantization continues after the model converges. 

\subsection{Implementation Details}

\subsubsection{Training Algorithm}
Our proposed method is shown in Algorithm~\ref{alg:training}. We prune the model with a ratio from $r_l=0.3$ to $r_h=0.7$ based on \cite{liu2018rethinking,zhu2016ttq}. The overall process can be summarized as:
i) Pre-training with hyperspherical learning architecture \cite{liu2017deephyperspherical}; ii) Iterative preprocessing the model weights, \textit{i.e.}, prune the model to target sparsity $r_h$ and reset the weights via $\mathbf{W}=\texttt{Ternary}(\mathbf{W},0)$ after each pruning (Fig.~\ref{fig:whole}); iii) Ternary quantization, updating the weights and $\Delta$ through STE. 
\begin{algorithm}
	\caption{HQ training approach}
	\label{alg:training}
	\begin{algorithmic}[1]
		\State \textbf{Input:} Input $\mathbf{x}$, a hyperspherical neural network $\phi(\mathbf{W},\cdot)$, $r_{l}=0.3$, $r_{h}=0.7$, and step size $\delta=0.01$.
		\State \textbf{Result:} Quantized ternary network for inference

        
        
        \State\textbf{1. Preprocessing:}
        \State $r = r_{l}$
        \While{ \label{lst:line:5}
		\textit{$r<r_{h}$}
		} \Comment{Iterative pruning and reinitializing }
        \State 
		$\mathbf{M}=\texttt{Mask}(\mathbf{W},r)$ \Comment{Obtain the pruning mask}
	    
	    \State
	    $\mathbf{W}=\mathbf{W}\odot\mathbf{M}$ \Comment{Pruning}
	    
	    \State $\mathbf{W}=\texttt{Ternary}(\mathbf{W},0)$  \Comment{Reinitialization }

		\While{
		\textit{not converged}
		} \label{lst:line:9}
		
		\State{
	    $y=\phi ((\mathbf{M}\odot\mathbf{W}), \mathbf{x})$ \Comment{Eq.~\eqref{eq:spop} }
	    }
	    \State{
	    Perform SGD, calculate $\frac{\partial L}{\partial {\mathbf{W}}}$, and update $\mathbf{W}$
	    }
		\EndWhile \label{lst:line:12}
		\State{$r+= \delta$} \Comment{Increase the pruning ratio $r$} \label{lst:line:13}
		\EndWhile \label{lst:line:13}

		\State\textbf{2. Ternary Quantization:}
		\While{\textit{not converged}} \label{lst:line:16}
		\State{
		$\mathbf{\hat{W}}=\texttt{Ternary}(\mathbf{W},\Delta)$ }
	    \State{
	    $y=\phi ((\mathbf{M}\odot{\mathbf{\hat{W}}}), \mathbf{x})$
	    }
	    \State{
	    Get $\frac{\partial L}{\partial {\mathbf{{W}}}} $ via SGD; update $\mathbf{{W}}$, $\Delta$ \Comment{Eq.~(\ref{eq9},\ref{eq19})}
	    }
		\EndWhile \label{lst:line:20}
	
	\end{algorithmic}
\end{algorithm}
\subsubsection{\textcolor{black}{Training Settings and Training Time}}
For image classification, the batch size is 128. The weight decay is 0.0001, and the momentum of stochastic gradient descent (SGD) is 0.9. We use the cosine annealing schedule with restarts every 10 epochs \cite{loshchilov2016sgdr} to adjust the learning rates. The initial learning rate is 0.001. All of the experiments use 16-bit half-precision from PyTorch to accelerate the training process. Thus, all parameters (except codebook with 2-bit) are stored as 16-bit precision values. 

It takes about 50 epochs to convert a Pytorch model to a hyperspherical one. The SGD loop (Line \ref{lst:line:5}-\ref{lst:line:13} in Algorithm \ref{alg:training}) takes at least 200 epochs.
The ternary quantization loop (Line \ref{lst:line:16}-\ref{lst:line:20} in Algorithm \ref{alg:training}) takes about 200 epochs. 

\subsubsection{\textcolor{black}{Compression Strategy}}
Inspired by vector quantization, we use codebook as the compression method \cite{stock2019killthebits,martinez_2020_pqf}. Unlike the vector quantization methods that use a learned codebook, we use the Huffman table \cite{van1976huffman} as a fixed codebook (Table 2 in the Appendix). We use gzip to finalize the model file as in other work \cite{prune_zip,stock2019killthebits}.
It is shown that the Huffman table can maximize the compression effect when each codeword consists of three ternary values, \textit{e.g.},  $\{0, 0, 0\}$. Note that the Huffman table can only boost the compression when the high-frequency patterns exist, such as $\{0,0,0\}$. For other works with low-sparsity or non-sparse quantization \cite{stock2019killthebits,cho2021dkm,martinez_2020_pqf,zhu2016ttq}, applying the Huffman table may not help compress the model. We find inconsistent compression results in ABGD \cite{stock2019killthebits} and PQF \cite{martinez_2020_pqf}. The actual size of their models (obtained from github in gzip format) is shown in Table \ref{tab_w2a32} is different from the results stated in their paper.

\section{Experiments}
Our experiments involve image classification and object detection tasks. We evaluate our method on ImageNet data set \cite{ILSVRC15} with MobileNetV2 \cite{sandler2018mobilenetv2} and ResNet-18/50 \cite{he2016deep}. For object detection, we use the MS COCO \cite{lin2014mscoco} dataset and Mask R-CNN \cite{he2017mask,wu2019detectron2}. The pre-trained weights are provided by the PyTorch zoo and Detectron2 \cite{wu2019detectron2}. 

\subsection{Image Classification}
\begin{figure*}
     \centering
     \begin{subfigure}[b]{0.3\textwidth}
         \centering
         \includegraphics[width=\textwidth]{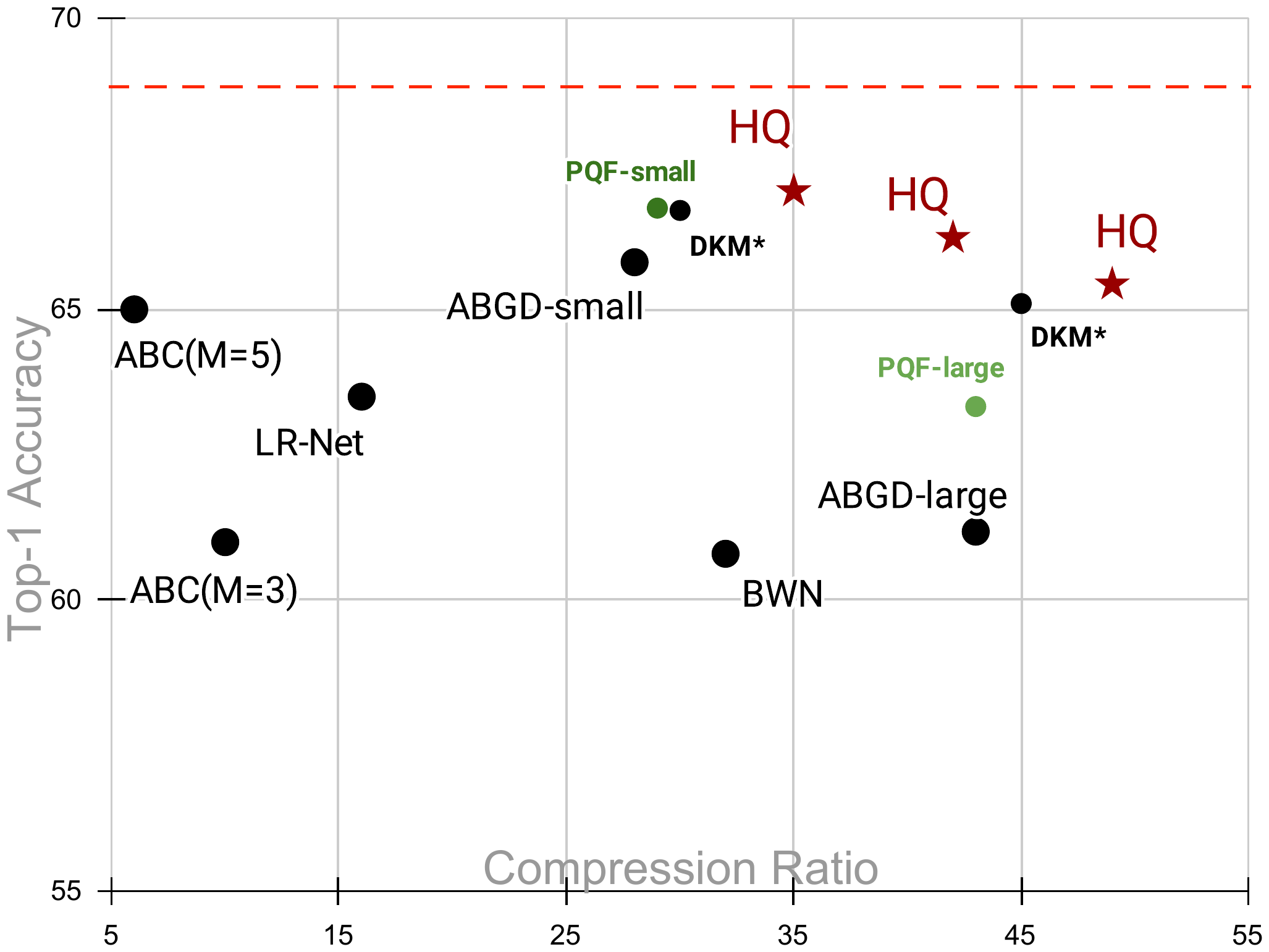}
         \caption{ResNet-18 on ImageNet.}
         \label{fig:y equals x}
     \end{subfigure}
     \begin{subfigure}[b]{0.3\textwidth}
         \centering
         \includegraphics[width=\textwidth]{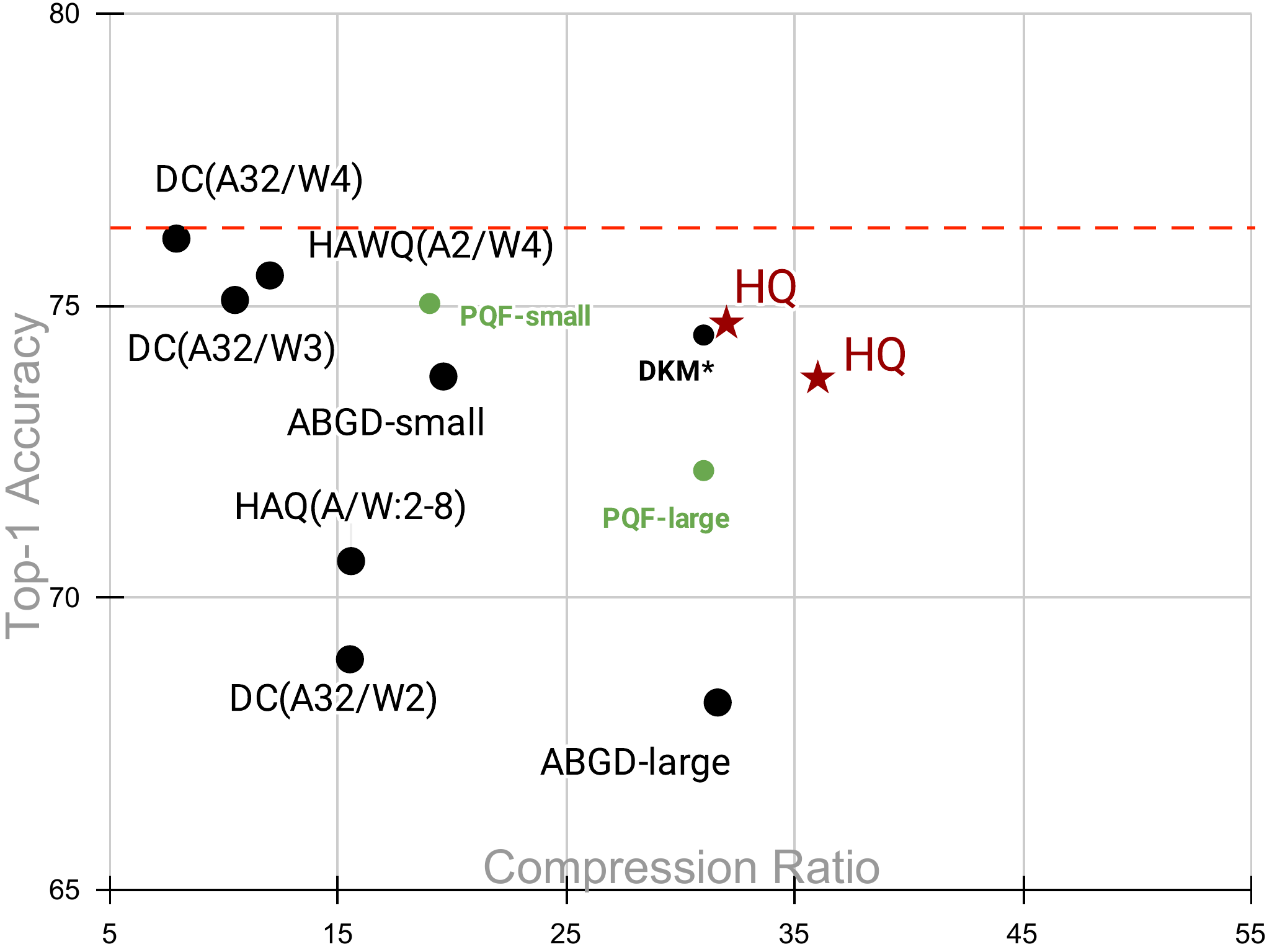}
         \caption{ResNet-50 on ImageNet.}
         \label{fig:three sin x}
     \end{subfigure}
    \caption{The compression ratio and accuracy of ResNet-18/50 on ImageNet. Our method achieves much higher accuracy and compression ratio compared to other work. The dash-line is the baseline accuracy from PyTorch zoo.}
    \label{img:modelsize}
\end{figure*}
\begin{table}[t]
    \footnotesize
     \caption{Ternary quantization results on ImageNet. We leave the last FC layer as full-precision like other works.} 
        \centering
        \begin{adjustbox}{max width=0.4\textwidth}
        \begin{tabular}{lllcc}
        \toprule
        \textbf{Models} &
        \textbf{Methods}
        & 
        \textbf{Acc.}
        \\
        \midrule
        \multirow{9}{6em}{ResNet-18\\Acc.: 69.76}& 
         \textsc{\textbf{HQ (Ours)}}   & \textbf{68.5} &   \\
         & \textsc{TWN (2016) \cite{li2016twn}} &   61.8  \\
        & \textsc{TTQ (2016) \cite{zhu2016ttq}} &  66.6 & \\
        & \textsc{INQ (2017) \cite{zhou2017incremental}} &   66.0  \\
        & \textsc{ADMM (2018) \cite{leng2018extremelyadmm}} &   67.0  \\
        & \textsc{LQ-Net (2018) \cite{zhang2018lqnet}} &   68.0  \\
        & \textsc{AdaRound (2020) \cite{nagel2020up_adaround}}   & 55.9\\  & \textsc{BRECQ (2021) \cite{li2021brecq}} &   66.3  \\
        & \textsc{RTN (2020) \cite{li2020rtn_23}} &   68.5  \\
        
        \midrule
        \multirow{6}{6em}{ResNet-50\\Acc.: 76.15}&  
         \textbf{HQ (Ours)}    & \textbf{75.2}  \\
         
        & \textsc{TWN\cite{li2016twn}}  & 72.5   \\
        & \textsc{LQ-Net \cite{zhang2018lqnet}} &   75.1 \\
        & \textsc{AdaRound \cite{nagel2020up_adaround}}    & 47.5\\  & \textsc{BRECQ \cite{li2021brecq}}    & 72.4  \\
        \bottomrule
        \end{tabular}
        \end{adjustbox}
   
    \label{tab_w2a32_0}
\end{table}

\begin{table*}[!htbp]
        \centering
        \caption{Model compression results on ImageNet. ``Bits (W/A)'' denotes the bit-width of weight and activation. ``Ratio'' denotes the storage compression level. ``+'' denotes 16-bit weight precision in FC layer. ``Size'' denotes disk size. ``*'' indicates gzip-compressed publicly available models. The detailed bit allocations can be found in the Appendix.} 
        \begin{adjustbox}{width=0.7\textwidth}
        \begin{tabular}{lclccccc}
        \toprule
        \textbf{Models} &
        \textbf{Comp. Level} &
        \textbf{Methods}
        & 
        \textbf{Bits \tiny(W/A)}
        & 
        \textbf{Acc.}
        & \textbf{Size} 
        & \textbf{Ratio}  
        \\
        \midrule
        \multirow{8}{6em}{ResNet-18\\Acc.: 69.76\\Size: 45 MB}&  \multirow{4}{4em}{$\sim$30$\times$} & 
         \textsc{\textbf{HQ (Ours)}} & \textbf{2/16}  & \textbf{67.03} & \textbf{1.23 MB} & \textbf{37$\times$} \\
         && \textsc{DKM (2022)\cite{cho2021dkm}} & 32/32  & 66.7 & 1.49 MB & 30$\times$   \\
        && \textsc{PQF* (2021)\cite{martinez_2020_pqf}} & 32/32  & 66.74 & 1.54 MB & 30$\times$  \\
        && \textsc{ABGD* (2020)\cite{stock2019killthebits}} & 32/32  & 65.81 & 1.51 MB & 30$\times$  \\ \cmidrule{2-8}
        &\multirow{4}{4em}{$\sim$40$\times$} &  \textsc{\textbf{HQ (Ours)}} & \textbf{2/16}  & \textbf{65.48} & \textbf{939 KB} & \textbf{48$\times$}    \\
        && \textsc{DKM(2022)\cite{cho2021dkm}} & 32/32  & 65.1 & 1 MB & 45$\times$&  \\
        &&  \textsc{PQF* (2021)\cite{martinez_2020_pqf}} & 32/32 & 63.33 & 1.04 MB & 43$\times$ \\
        && \textsc{ABGD* (2020)\cite{stock2019killthebits}} & 32/32  & 61.18 & 1.01 MB & 45$\times$  \\
        \midrule
        \multirow{13}{6em}{ResNet-50\\Acc.: 76.15\\Size: 99 MB}&  \multirow{7}{3em}{10$-$20$\times$} & 
         \textbf{HQ (Ours)} & \textbf{2/16}$^+$  & \textbf{75.2} & 6.89 MB & 14$\times$   \\
         &&\textsc{HAWQ (2019)\cite{dong2019hawq}} & {2$\sim$8/4$\sim$8}  & {75.4} & 7.96 MB & 12$\times$  \\
        && \textsc{PQF* (2021)\cite{martinez_2020_pqf}} & 32/32  & 75.04 & 5.10 MB & 19$\times$   \\
        &&\textsc{TQNE (2020)\cite{fan2020training}} & 32/32  & 74.3 & - & 19$\times$\\
        &&\textsc{ABGD* (2020)\cite{stock2019killthebits}} & 32/32  & 73.79 & 5.01 MB & 20$\times$  \\
        &&\textsc{HAQ (2019)\cite{wang2019haq}} &2$\sim$8/2$\sim$8 &	70.63&6.30MB &16$\times$ \\
        && \textsc{DC (2015)\cite{han2015deep}} & 2/32& 68.95 &6.32MB & 16$\times$\\
        
         \cmidrule{2-8}
        &\multirow{6}{3em}{$\sim$30$\times$} &  \textsc{\textbf{HQ (Ours)}} & \textbf{2/16}  & \textbf{73.87} & \textbf{2.63 MB} & \textbf{38$\times$}   \\
        &&\textsc{\textbf{HQ (Ours)}} & \textbf{2/16}&	\textbf{74.7}&	\textbf{3.01MB} &\textbf{33$\times$} \\
        && \textsc{DKM (2022)\cite{cho2021dkm}} & 32/32  & 74.5 & 3.32 MB & 29$\times$ \\
        &&  \textsc{PQF* (2021)\cite{martinez_2020_pqf}} & 32/32 & 72.18 & 3.26 MB & 30$\times$ \\
         && \textsc{TQNE (2020)\cite{fan2020training}} & 32/32  & 68.8 & - & 32$\times$  \\
        && \textsc{ABGD* (2020)\cite{stock2019killthebits}} & 32/32  & 68.21 & 3.16 MB & 31$\times$  \\
        \midrule
        \multirow{4}{6em}{MobileNetV2\\Acc.: 71.88\\Size: 14 MB} &\multirow{4}{3em}{15$-$20$\times$} 
        & \textsc{\textbf{HQ (Ours)}} & 2/16  & 58.74 & \textbf{0.71 MB} & \textbf{20$\times$}\\
        && \textsc{HAQ (2019)\cite{wang2019haq}} & {2$\sim$8/2$\sim$8} & {66.75} & 0.95 MB & 15$\times$\\
        && \textsc{BRECQ (2021)\cite{li2021brecq}} & 2/8  & 56.29 & 0.83 MB & 17$\times$ \\
        && \textsc{Han \textit{et al.} \cite{han2015deep,wang2019haq} (2015)} & 2/32  & 58.07 & 0.96 MB & 17$\times$\\
        \bottomrule
        \end{tabular}
        \end{adjustbox}
    \label{tab_w2a32}
\end{table*} 

Following the practices of mainstream model compression work \cite{stock2019killthebits,dong2019hawq,martinez_2020_pqf}, when compressing the model size, we quantize all of the weights of the convolution and the fully-connected (FC) layers to 2-bit (except the first layer). We compare our results with leading compression results from PQF~\cite{martinez_2020_pqf}, ABGD~\cite{stock2019killthebits}, BRECQ~\cite{li2021brecq}, HAWQ~\cite{dong2019hawq}, and TQNE~\cite{fan2020training}. We also compare our work with milestone approaches, such as ABC-Net~\cite{lin2017abcnet}, Deep Compression (DC)~\cite{han2015deep}, Hardware-Aware Automated Quantization (HAQ)~\cite{wang2019haq}, Hessian AWare Quantization (HAWQ)~\cite{dong2019hawq}, LR-Net~\cite{shayer2017learning}, and BWN~\cite{rastegari2016xnor}. 

Our method significantly outperforms leading model compression methods in terms of bit-width, compression ratio and accuracy (Fig.~\ref{img:modelsize}, Table~\ref{tab_w2a32}).
It compresses ResNet-18 from 45 MB to 1.28MB (35$\times$ compressed) while maintaining high accuracy (67.03\% vs. 69.7\% of the original model), and compress ResNet-50 from 99 MB to 3.1MB (32$\times$ compressed) with an accuracy of 74.7\% (vs. 76.15\% of the original model). In the extreme cases, our method produces much smaller models with less than 4\% accuracy drop, for example, 935KB out of 45 MB  (48$\times$ compressed) for ResNet-18 and 2.6 MB out of 99 MB  (37$\times$ compressed) for ResNet-50. Although DKM's results \cite{cho2021dkm} are close to ours, their models cannot  reduce the memory footprint.

For MobileNetV2, our method performs better at above 15$\times$ compression level. Quantizing the pointwise layer of MobileNet leads to significant accuracy loss \cite{gope2020ternary_mob}. That is the reason why the fully quantized 2-bit precision methods \cite{han2015deep,li2021brecq} have low accuracy. 
Whereas, HAQ \cite{wang2019haq} applies mixed-precision to improve accuracy. In addition, the works \cite{dbouk2020dbq_mob,gope2020ternary_mob,sung2015resiliency_mob} show that a less complex model is sensitive to ternary quantization due to the potential lack of redundant representation capability. 

We also compare our work with conventional ternary quantization works which leave the last FC layer as full-precision (Table \ref{tab_w2a32_0}). Our work achieves comparable results with other leading methods.

\subsection{Object Detection and Segmentation}
Similar to previous work~\cite{stock2019killthebits,martinez_2020_pqf,li2021brecq}, we test our method on the Mask R-CNN \cite{he2017mask} architecture with ResNet-50 backbone to verify its generalizability. The source code, hyperparameters and the pre-trained model are provided by Detectron2~\cite{wu2019detectron2}. We apply our method to the entire model except for the first layer. We compare against recent baselines, such as the ABGD~\cite{stock2019killthebits}, PQF~\cite{martinez_2020_pqf}, and BRECQ~\cite{li2021brecq}. As shown in Table~\ref{tab_coco_quant}, compared to ABGD and PQF, our method gives a higher compression ratio and a similar or better recognition result.


\begin{table}[t]
    \footnotesize
    	\caption{The model size and Average Precision (AP) with bounding box (bb) and mask (mk) are compared. 
	}
        \centering
        \begin{adjustbox}{max width=0.4\textwidth}
        \begin{tabular}{cccccc}
        \toprule
         \textbf{Methods} & \textbf{ Bits}\tiny\textbf{ (W/A)} & \textbf{AP$^{bb}$}&\textbf{AP$^{mk}$}  & \textbf{Size}& \textbf{Ratio} \\
        \midrule
        \textsc{Baseline} &32/32&{37.9}&{34.6}&{170MB}&1$\times$\\
        \midrule
         \textsc{\textbf{HQ (Ours)}} &2/16&{35.0}&{31.7}&\textbf{4.92MB}&\textbf{34}$\times$\\ 
         \textsc{ABGD (2020)} &32/32&33.9&30.8&6.6MB&26$\times$\\
         \textsc{PQF (2021)} &32/32&36.3 &33.5 &6.6MB &26$\times$\\
         \textsc{BRECQ (2021)} &2/8&34.23&- &-&-\\
        
        \bottomrule
        \end{tabular}
        \end{adjustbox}
        
        \label{tab_coco_quant}
    \end{table}   
\subsection{\textcolor{black}{Model Size and Accuracy}}

\begin{figure}[H]
\centering
\includegraphics[width=0.7\columnwidth]{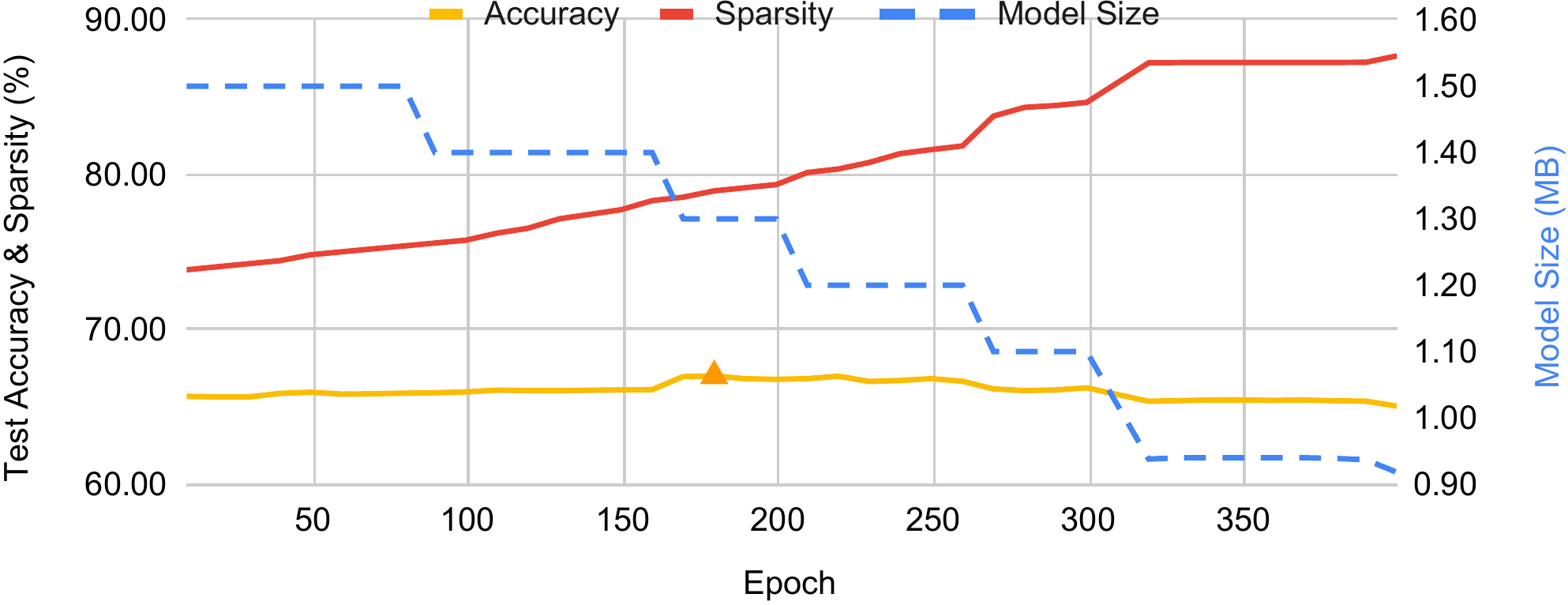} 
\caption{
The trend lines of size-accuracy and sparsity of ResNet-18. The blue dashed line denotes the model size. 
}
\label{fig:sa_ab}
\end{figure}

The accuracy, sparsity and size of the quantized ResNet-18 models are shown in Table~\ref{size_acc_ab} and Fig.~\ref{fig:sa_ab}. The results show that the model accuracy will increase with the sparsity until it reaches a certain level (the triangle symbol in Fig.~\ref{fig:sa_ab}). Then the accuracy starts to decrease as the pruning continues, which is the same as the Figure 5 in TTQ \cite{zhu2016ttq}. This phenomenon is different from pruning, where the accuracy decreases linearly as the sparsity increases. One possible reason is that the capacity \cite{sung2015resiliency_mob} of the quantized model changes with the portion of $0$ and $\pm\frac{1}{\sqrt{|\mathbf{I}_\Delta|}}$. The model capacity is low when the number of $\pm\frac{1}{\sqrt{|\mathbf{I}_\Delta|}}$ is dominant (close to binary). As the sparsity increases, the weight tends to become ternary, whose capacity is higher than binary weights. Pruning does not have this issue since it has full-precision weight values. As the proportion of $0$ keeps increasing, the capacity will drop, leading to accuracy drop. 
\begin{table}[H]
      \footnotesize
       
        \caption{Size-accuracy results of ResNet-18 on ImageNet. 
	}
        \centering

        \begin{adjustbox}{max width=0.4\textwidth}
        \begin{tabular}{llcrrrrr}
        \toprule
        
         \textbf{Epoch} &  \textbf{Accuracy (\%)} &  \textbf{Sparsity (\%)} &  \textbf{Size} \\ 
         \midrule
        19 & 65.64 & 74.06 & 1.50MB \\ 
        99 & 65.97 & 75.77 & 1.40MB \\ 
        179 & 67.03 & 78.96 & 1.30MB \\ 
        259 & 66.66 & 81.86 & 1.20MB \\ 
        299 & 66.23 & 84.67 & 1.10MB \\ 
        389 & 65.37 & 87.26 & 0.94MB \\ 
    \bottomrule
        \end{tabular}
        \end{adjustbox}
        	
        \label{size_acc_ab}
    \vspace{-2mm}
    \end{table}
    
\section{\textcolor{black}{Ablation Study}}
\label{abst}
We experiment with the criteria including different pruning settings, reinitialization, and hyperspherical learning. ``\textsc{\textbf{Hyper}}'' means hyperspherical training. ``\textsc{\textbf{Pruning+Reinit}}'' means pruning with reinitialization. ``\textsc{\textbf{Baseline/BL}}'' means the pre-trained model from PyTorch or Detectron\cite{wu2019detectron2}. Experiments in  the Section \ref{htq}
and \ref{prset} apply full-precision to the last FC layer.

\subsection{Hyperspherical Pruning and Reinitialization}
We study the change of the cosine distance $\mathcal{D}$ (Eq.~(\ref{cosdist})) brought by our proposed method:  applying hyperspherical preprocessing method prior to ternary quantization. 

Table~\ref{tab_res_quant_distance} shows that pruning can reduce $\mathcal{D}$ with or without hyperspherical learning and $\mathcal{D}$ tends to decrease along with the growing sparsity. ``\textsc{\textbf{Pruning+Reinit}}'' significantly enlarges the distance gap with the pruning-only results and can improve model's performance \cite{zhou2019deconstructing,frankle2018lottery}. 

Table~\ref{tab_coco_quant_distance} shows the difference between applying reinitialization once (\textsc{\textbf{Reinit}$_1$}) and 10 times (\textsc{\textbf{Reinit}$_{10}$}) to the target sparsity of 0.8. 
``\textsc{\textbf{Reinit}$_{10}$}'' has a smaller average distance than ``\textsc{\textbf{Reinit}$_{1}$}'', indicating that iterative pruning and reinitialization encourage model weights close to its ternary version (Section \ref{increase_s}, Fig.~\ref{fig:sp_nfix}). 

\begin{table}[t]
    \footnotesize
    	\caption{The cosine distance $\mathcal{D}$ of convolutional layers of ResNet-50 on ImageNet.  
	}
        \centering
        \begin{adjustbox}{max width=0.4\textwidth}
        \begin{tabular}{ccccc}
        \toprule
         \textbf{Methods} & \textbf{$\mathcal{D}$}  & \textbf{Accuracy}&  \tiny{\textbf{Sparsity}}  \\
        
        \midrule
        \textsc{Baseline} &0.262&{76.15}&0.0\\
        \textsc{Hyper} &0.288&{76.18}&0.0\\
        \midrule
         \textsc{Baseline+Pruning} &0.187&{76.06}&0.4\\ 
         \textsc{Hyper+Pruning} &0.155&76.99&0.4\\
         \textsc{Hyper+Pruning+Reinit} &\textbf{0.068}&77.04 &0.4\\
         \midrule
         \textsc{Baseline+Pruning} &0.149&76.09&0.6\\ 

         \textsc{Hyper+Pruning+Reinit} &\textbf{0.056}&77.03&0.6\\ 
        
        \bottomrule
        \end{tabular}
        \end{adjustbox}
        
        \label{tab_res_quant_distance}
    \end{table}

    \begin{table}[t]
    \footnotesize
    	\caption{Distance comparison of convolutional layers on the object detection task. 	}
       
        \centering
        \begin{adjustbox}{max width=0.45\textwidth}
        \begin{tabular}{cccccc}
        \toprule
         \textbf{Methods} & \textbf{$\mathcal{D}$}  & \textbf{AP$^{bb}$}& \textbf{AP$^{mk}$} & \tiny{\textbf{Sparsity}}  \\
         
        \midrule
        \textsc{Baseline} &0.263&{41.0}&{37.2}&0.0\\
        \textsc{Hyper} &0.246&{41.03}&{37.54}&0.0\\ 
        \midrule
         
         \textsc{Hyper+Pruning} &0.193&41.33&37.94&0.8\\
         \textsc{Hyper+Pruning+Reinit}$_1$ &\textbf{0.186}&40.92 &37.33&0.8\\
         \textsc{Hyper+Pruning+Reinit}$_{10}$ &\textbf{0.113}&40.31 &36.86&0.8\\

        
        \bottomrule
        \end{tabular}
        
        \end{adjustbox}
         \label{tab_coco_quant_distance}
    \end{table}
\subsection{Hyperspherical Ternary Quantization}
\label{htq}
We perform ternary quantization on ResNet-18 to examine the impact of hyperspherical learning and ``\textsc{\textbf{Pruning+Reinit}}''.
The models are initialized by three different pre-trained weights (Table \ref{tab:abhter}). 
The initialized models are quantized with hyperspherical learning and regular training (``\textsc{\textbf{Non-Hyper}}'') by 100 epochs. Table \ref{tab:abhter} shows significant improvements brought by hyperspherical learning and ``\textsc{\textbf{Pruning+Reinit}}''. The figure of trend lines can be found in the Appendix.

\begin{table}[t]
    \footnotesize
    \caption{Quantization accuracy of ResNet-18 on ImageNet.}
        \centering
        \begin{adjustbox}{max width=0.4\textwidth}
        \begin{tabular}{lll}
    \toprule
        \textbf{Initial Models (Accuracy)} & \textsc{\textbf{Hyper}} & \textsc{\textbf{Non-Hyper}} \\
    \midrule
        \textsc{Baseline} (69.76) & 66.45 & 60.46 \\  
        \textsc{BL+Pruning+Reinit} (69.63) & 67.17 &  66.11\\  
        \textsc{\textbf{Hyper}+BL+Pruning+Reinit} (69.67) & 67.50 & 65.24 \\  
    \bottomrule
    \end{tabular}
    \end{adjustbox}
    \label{tab:abhter}
\end{table}
\subsection{The impact of Pruning Settings}
\label{prset}
\begin{figure}[H]
\centering
\includegraphics[width=0.7\columnwidth]{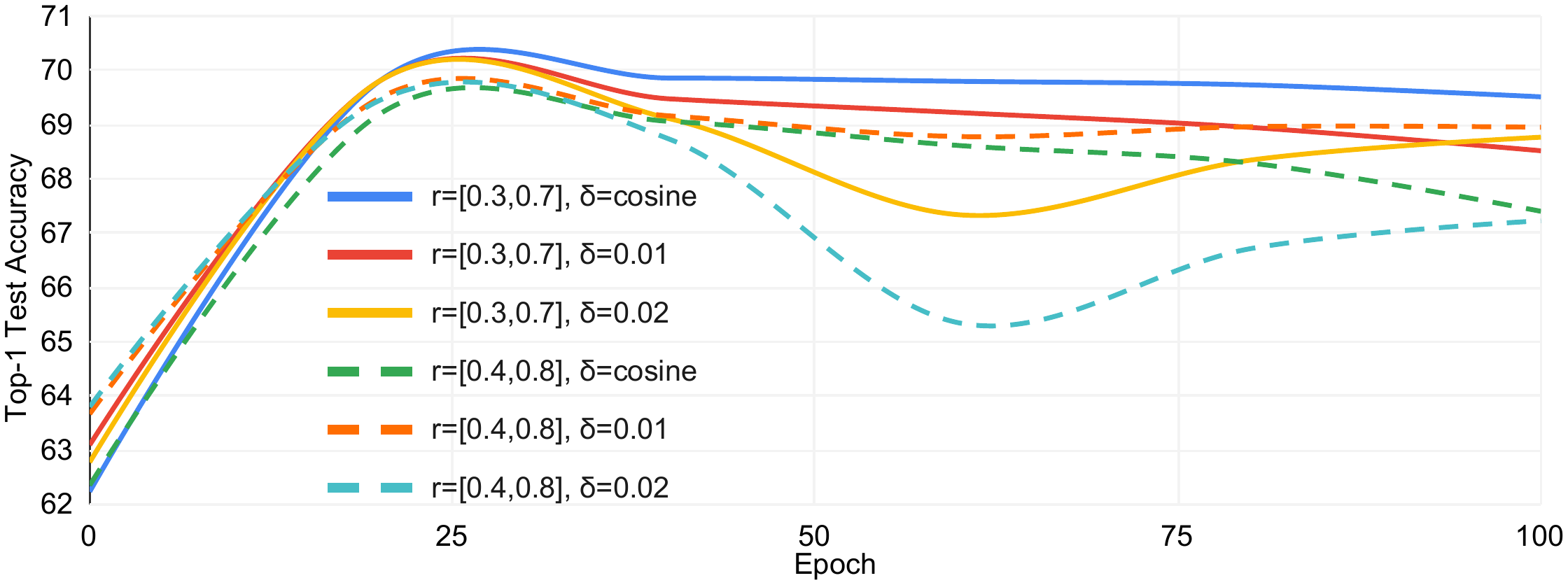} 
\caption{
The accuracy trends of ResNet-18 with different pruning settings during preprocessing. 
}
\label{fig:sa_ab_0}
\end{figure}

\begin{figure}[H]
\centering
\includegraphics[width=0.7\columnwidth]{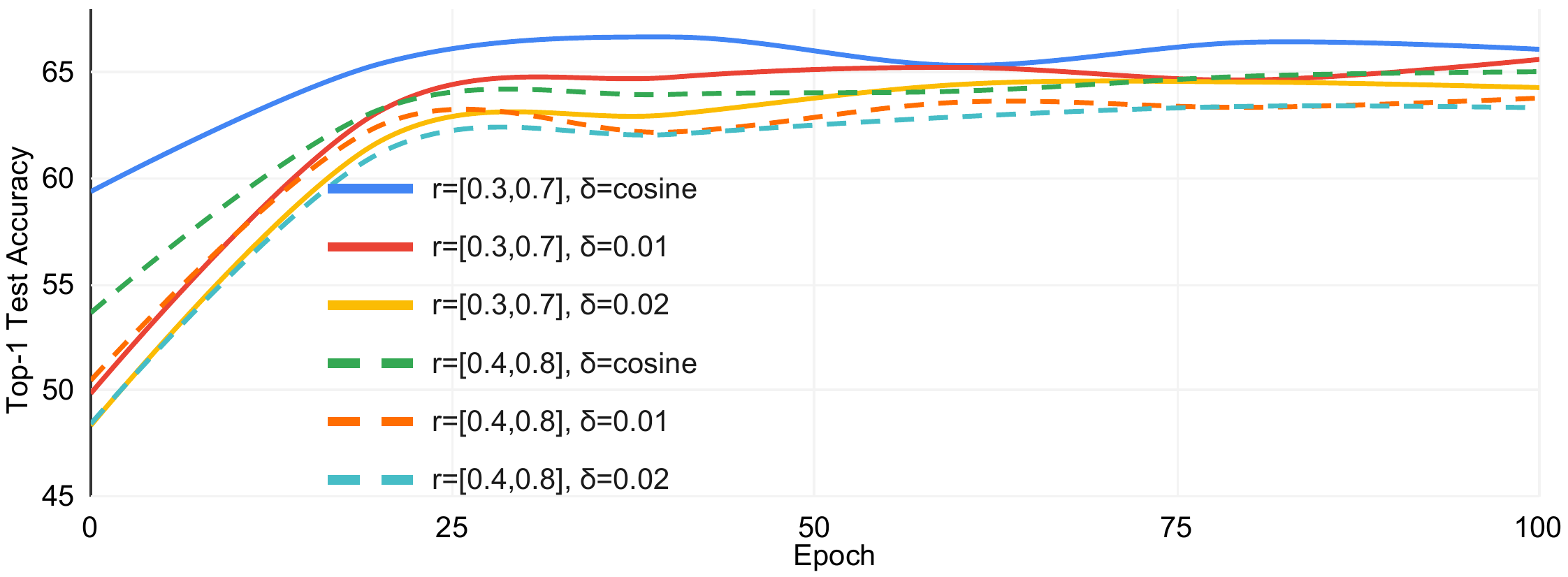} 
\caption{
The following ternary quantization accuracy of ResNet-18 after preprocessing. 
}
\label{fig:sa_ab_1}
\end{figure}

We further study the impact of different pruning ranges ($r$) and step sizes ($\delta$, line \ref{lst:line:13} of Algorithm~\ref{alg:training}) on ResNet18. The Figure 5 of TTQ \cite{zhu2016ttq} indicates that a proper pruning range before ternary quantization is from 0.3 to 0.7, and we take that as a reference.
During preprocessing, $r$ changes from $0.3$ to $0.7$ or $0.4$ to $0.8$. The step sizes can be $0.01$, $0.02$, or controlled by cosine annealing. The ``\textsc{\textbf{Pruning+Reinit}}'' starts at the 20-th epoch. The total training epochs are 100. Fig.~\ref{fig:sa_ab_0} shows that pruned models with larger step sizes and ratios perform poorly and take longer to recover. Using cosine annealing \cite{loshchilov2016sgdr} method to adjust the step size $\delta$ improves the overall performance. Fig.~\ref{fig:sa_ab_1} shows the following ternary quantization results. Cosine annealing accelerates the convergence. 
\section{Conclusion}
We propose a novel method, Hyperspherical Quantization, to construct sparse ternary weights by unifying pruning, reinitialization and ternary quantization on the hypersphere. The proposed iterative pruning and reinitialization strategy greatly outperform state-of-the-art model compression results in terms of size-accuracy trade-offs. A major contribution of our method is the use of hyperspherical learning to enhance the compression capability. Our work further reveals and demonstrates that pruning and quantization are linked through hypersphere. Our work also explores a new way to extremely compress DNN models without using clustering. Future work may combine our method with ternary activation quantization~\cite{chen2020fatnn,li2019apot} to further speed up the inference.
{\small
\bibliographystyle{ieee_fullname}
\bibliography{egbib}
}

\end{document}